\documentclass[runningheads]{llncs}
\usepackage{graphicx}
\usepackage{hyperref}
\usepackage[utf8]{inputenc}
\usepackage[english, russian]{babel}

\usepackage{graphicx}
\graphicspath{ {images/} }

\usepackage{caption} 
\usepackage{indentfirst}

\usepackage{amsmath}
\usepackage{amsbsy} % Bold math

\usepackage{multirow}

\begin{document}

\selectlanguage{english}

\title{RuREBus: a Case Study of Joint Named Entity Recognition and Relation Extraction from e-Government Domain\thanks{The extended notes for invited talk "When CoNLL-2003 is not Enough: are Academic NER and RE Corpora Well-Suited to Represent Real-World Scenarios?" delivered by Ivan Smurov}}
\titlerunning{RuREBus}
 \author{ 
Vitaly Ivanin \inst{1,2}\orcidID{0000-0002-4128-6151} \and
Ekaterina Artemova \inst{3}\orcidID{0000-0003-4920-1623} \and
Tatiana Batura \inst{4,7}\orcidID{0000-0003-4333-7888} \and
Vladimir Ivanov\inst{5,7} \orcidID{0000-0003-3289-8188} \and
Veronika Sarkisyan \inst{3} \orcidID{0000-0001-5681-7632} 
\and 
Elena Tutubalina\inst{6,7}\orcidID{0000-0001-7936-0284} \and
Ivan Smurov\inst{1,2} \orcidID{0000-0001-8937-8883}
}
% %
 \authorrunning{Ivanin et al.}
 \institute{
 ABBYY
 \and
 Moscow Institute of Physics and Technology 
 \and
National Research University Higher School of Economics, Moscow, Russia
\and
Novosibirsk State University
\and 
Innopolis University
\and 
Kazan Federal University
\and 
Lomonosov Moscow State University\\
\email{ivan.smurov@abbyy.com}
}

\maketitle              % typeset the header of the contribution
\begin{abstract}

We show-case an application of information extraction methods, such as named entity recognition (NER) and relation extraction (RE) to a novel corpus, consisting of documents, issued by a state agency. The main challenges of this corpus are: 1) the annotation scheme differs greatly from the one used for the general domain corpora, and 2) the documents are written in a language other than English. Unlike expectations, the state-of-the-art transformer-based models show modest performance for both tasks, either when approached sequentially, or in an end-to-end fashion.  Our experiments have demonstrated that fine-tuning on a large unlabeled corpora does not automatically yield significant improvement and thus we may conclude that more sophisticated strategies of leveraging  unlabelled  texts are demanded.
In this paper, we describe the whole developed pipeline, starting from text annotation, baseline development, and designing a shared task in hopes of improving the baseline. Eventually, we realize that the current NER and RE technologies are far from being mature and do not overcome so far challenges like ours.

\end{abstract}

\section{Introduction} % vanya, katya, tanya 
% merge with related work for the sake of space 

% здесь пытаюсь рассказать о том, что существующиетыатасеиы слишком академичны и оценки, полученные на них для каких-то моделей сильно завышены по сравнению с практикой (КА)
Information extraction tasks, named entity recognition (NER) and relation extraction (RE), have been studied extensively. NER and RE are sometimes thought of as easy and almost solved problems. However, outside of the idealistic academic setup, many complications may arise. The most used datasets, leveraged to compare new methods and establish state-of-the-art (SoTA) results are CoNLL03 \cite{conll03}, TACRED \cite{zhang2017tacred}, SemEval-2010 Task 8 \cite{hendrickx-etal-2010-semeval}, CoNLL04 \cite{carreras-marquez-2004-introduction}, ACE 2005 \cite{Walker}, OntoNotes \cite{10.5555/1614049.1614064}. However, as the choice of open sources for dataset construction appears to be quite limited, these datasets are usually assembled from news articles. What is more, the annotation scheme typically is driven by academic interest, rather than practical considerations. Real-life applications though may vary a lot and target domains other than news. Such applications cover Legal Tech (previous studies had focused on extraction of organization and person names \cite{dozier2010named}, while more recent studies look beyond classical NER types \cite{cardellino2017low,teruel2018legal,leitner2019fine,leitner2020dataset}, medical domain \cite{huang2015community,weber2020huner} list more than forty corpora) and noisy user texts \cite{strauss2016results}, and may require a domain-specific and application-driven annotation scheme. It might prove to be difficult in practice to adopt exciting approaches to NER and RE to other domains, as straightforward domain adaption techniques do not lead to the desired quality. A question of how big the gap between academic benchmarks and real-life applications is rarely explored.

% здесь пытаюсь рассказать о том, что оценить применимость SoTA методов к другим языками сложно, тк может не быть SoTA датасетов 
Adaptation to languages other than English complicates the usage of the SoTA methods. If no corpora, similar to the one developed for English, in terms of size, domain, and annotation quality, is available, it is almost impossible to draw a fair comparison. For example, for the Russian language, used in this paper, the only corpora for joint NER and RE are FactRuEval \cite{FactRuEval2016},  significantly smaller than OntoNotes or TACRED, and RuRED \cite{rured}, which partially replicates TACRED annotation.
 As no identical setup for evaluation of NER and RE methods is available for different languages, it may be difficult to investigate whether the same methods deliver comparable results for different languages. Transfer learning \cite{yang2017transfer} is a promising paradigm that helps to re-use cross-lingual models, trained for English, for other languages. However, early attempts show that the application of transfer learning techniques turns out to be rather challenging. For example, so far, neither NER nor RE tasks benefit from transfer learning approaches, when applied to TACRED and RuRED.

In this paper, we explore a typical industrial case: prototyping NER and RE models in a specific application domain based on existing SoTA approaches. We describe a problematic real-life setup, which requires both 1) adaptation to a new domain and an unconventional annotation scheme and 2) processing text, written in a language other than English. Our results show a significant decrease in quality when compared to SoTA academic results. We aim to bring more attention to the challenges of NER and RE tasks and show that existing methods so far can be treated as off the shelf solutions only in a limited scope. 
The task under consideration comes from e-Goverment domain: we investigate the corpus of strategic planning documents\footnote{The corpus is open and available online on the Ministry of Economic Development of the Russian Federation website.}, which are annually issued by the Ministry of Economic Development of the Russian Federation. The entities considered relate to different types of state assets and enterprises. At the same time, the relations express various aspects of strategic planning, i.e., goal setting and forecasting. 
As the current approach to strategic planning is in desperate need of innovative organizational development, the NER and RE methods should be at the forefront of automation efforts. Extracted entities and relations between them allow for faster retrieval, ontology-based analysis, and compliance testing. 

The remainder is organized as follows. Section~\ref{sec:annotation} introduces the corpus and the annotation scheme. Section~\ref{sec:methods} presents with the methods used for NER and RE as well as with a shared task, which was held in hopes of improving baseline solution quality. Section~\ref{sec:conclusion} concludes by discussing the results and outlining the directions for future work.

\section{Corpus annotation} \label{sec:annotation} % vitaly (из DTSG)

% \begin{figure}[t]
%     \centering
%     \includegraphics[width=\textwidth]{img/brat.png}
%     \caption{Annotation interface for assigning entities and relations.}
%     \label{fig:brat}
% \end{figure}

We develop guidelines for entity and relation identification in order to maintain uniformity of annotation in our corpus. 

%Figure \ref{fig:brat} presents annotation interface for assigning entities and relations. 

\begin{table}
	\captionsetup{justification=centering}
    \caption{Entities description and examples (translated to English)}
    \label{tab:entities}

    \begin{tabular}{p{3cm}|p{4cm}|p{4cm}}
    Entity & Description  & Examples \\ \hline
    \texttt{MET} (metric) &  indicator or object on which the comparison operation is defined &   unemployment rate,  total length of roads, average life expectancy   \\ \hline
     \texttt{ECO} (economics) & economic entity  or infrastructure object  &   private business, PJSC Sberbank, hospital complex \\ \hline
      \texttt{BIN} (binary) & binary characteristics or single action   &  modernization, development, invest  \\ \hline
      \texttt{CMP} (comparative) & comparative characteristic  &    reduction of level, positive dynamics, increase of \\ \hline
      \texttt{QUA} (qualitative) & quality characteristic  &  ineffective, fault tolerant, stable   \\ \hline
      \texttt{ACT} (activity) & activities, events or measures  taken by the authorities  &  restoration work, educational project  ``Silver University'' , drug prevention   \\ \hline
      \texttt{INST} (institutions) &  institutions, structures   and organizations& 
        cultural center,  region administration ,  youth employment center  \\ \hline
      \texttt{SOC} (social) &  social object &  leisure activities, historical heritage,   population of the country   \\ 
    \end{tabular}

\end{table}
We define eight types of entities described in Table \ref{tab:entities} and nine types of relations that describe actions taken in the past and present time and also forecasts. We distinguish them in terms of tonality, whether the actions or state of affairs are positive, negative, or neutral. Another two relations are \texttt{GOL}, used for abstract goals, and  \texttt{TSK} used for specific tasks. Such relations tightly correspond to the domain: strategic planning is based on setting goals and targets due to past, current, and predicted state of affairs.

%\subsection{Annotation Process} %  

All annotations were obtained using a Brat Rapid Annotation Tool (BRAT) \cite{stenetorp2012brat}. Annotation instructions are available at the \href{https://github.com/dialogue-evaluation/RuREBus}{GitHub repository}\footnote{\url{https://github.com/dialogue-evaluation/RuREBus/}}. Each document in the corpus was annotated by two annotators independently, while a moderator resolved disagreements. To speed up and facilitate the annotation process, we used active learning techniques \cite{shen2017deep}. We applied widely used architecture, namely char-CNN-BiLSTM-CRF described in \cite{lample2016neural} and \cite{ma2016end} and used pre-trained FastText embeddings \cite{bojanowski2017enriching} from RusVectores \cite{KutuzovKuzmenko2017}. For RE, we employed morphological, syntactical, and semantic features obtained from
Compreno \cite{anisimovich2012,ZuyevK2013StatiSticalMT} and some hand-made features, such as capitalization templates and dependency tree distance between relation members.

The resulting corpus {\bf contains} 394,966 tokens, 120,989 entities and 12,648 relations. Annotation consistency is evaluated by measuring annotators {\bf agreement} on documents, which were marked up twice, leading to Cohen's kappa equal to 0.698.

% \paragraph{Corpus Analysis} %katya, vanya
% Entities/ relations
% POS tags?

% \begin{table}[]
%     \centering
%     \begin{tabular}{l|c|c|c|c|c|c|c|c|c}
%  Entity&ACT & BIN & CMP & ECO & INST & MET & QUA & SOC&Total\\ 
% \hline
% Total&12,274 &30,201  & 9,288  & 32,756 &3,756 & 14,161 &7,719& 10,834 & 120,989 \\
%  \end{tabular}
%     \caption{ Statistics of annotated entities }
%     \label{tab:e_stat}
% \end{table}

% \begin{table}[]
%     \centering
%     \begin{tabular}{l|c|c|c|c|c|c|c|c |c|c|c|c}
%  Relation&PPS & PNG & PNT & NPS & NNG & NNT & FPS & FNG& FNT& TSK& GOL&Total\\ 
% \hline
% Total&528&84 & 190  & 755 &844 &534 &1,167&229 & 141& 4,613& 3,563& 12,648 \\
%  \end{tabular}
%     \caption{ Statistics of annotated relations }
%     \label{tab:re_stat}
% \end{table}

\section{Baselines and Evaluation}  \label{sec:methods}

As a baseline for the NER task, we employed standard  BERT-based \cite{devlin2019bert} architecture, fine-tuned for the Russian language, namely  RuBERT \cite{kuratov2019adaptation} with an MLP on top of it. Although being close to SoTA on most academic corpora, this model yielded a rather disappointing strict token-based f1-score of \textbf{0.53}. 

To explore our corpus and double-check ourselves, we decided to conduct an external evaluation of both NER and RE tasks. As for our internal evaluation, we have chosen to continue working with NER (leaving the RE models exploration to the external evaluation).
External evaluation, organized in the form of RuREBus-2020 Shared Task, is vital in broadening the scope of tested models and providing additional validation for the scores obtained on the corpus. To provide an additional grounding for our results, we were also able to draw some comparisons between our setting and some other non-classical sequence-labelling tasks. 

\subsection{External evaluation} \label{sec:eval_extern}

We provide a full account on the RuREBus Shared Task results in \cite{rurebus}. Here we publish only the ones most useful for further analysis of the corpus.  

Unsurprisingly the most fruitful approach in the NER task was based on contextualised word embeddings in particular on BERT. While some participants attempted to use some additional layers such as BLSTMs and CRFs on top of contextualized word embeddings, the two systems with highest scores  both employed standard, but powerful MLP on top of BERT model. The scores obtained by the two systems are \textbf{0.561} and \textbf{0.547}\footnote{Obtained after the shared task deadline}. There is no significant difference between both models other than the version of BERT the participant used (multilingual uncased base BERT in the first case and RuBERT for the second one). We should also note that both systems only fitted BERTs on the train set and did not employ the finetuning on the 299M token unmarked corpus provided.

RE task yielded diverse models. Two top models, while once again both used BERT, had different architectures. One of the systems employed R-BERT \cite{wu2019enriching} based solution and was able to obtain 0.441 on RE task (given gold standard NERs). Another system used a SpanBERT \cite{joshi2020spanbert} inspired model. While two systems have substantial differences, scores obtained by them are roughly comparable (\textbf{0.441} for R-BERT and \textbf{0.394} for SpanBERT).

\subsection{Internal evaluation}

As a part of our internal evaluation, we have decided to fine-tune the language model of the contextual encoders for the NER task on the 299M token unmarked corpus.

In this ongoing work so far, we have been able to obtain some rather unexpected results. Fine-tuning our BERT-based baseline model did not leave to a significant improvement the performance and scored \textbf{0.54} on the NER task. We also tried fine-tuning multilingual BERT, but it scored only \textbf{0.44} on the NER task. In contrast fine-tuning ELMo \cite{peters2018deep} yielded the absolute best score obtained in both internal and external evaluation: \textbf{0.57}. We should also note that the ELMo model used for fine-tuning was pre-trained on English. Thus it had essentially non-random weights only on ``middle layers'' (as both embeddings and softmax were pre-trained on different vocabulary). 
We intend to explore this unexpected result further.

\subsection{Evaluation analysis}

During the internal and external evaluation, several SoTA-like models were tested, scoring \textbf{0.53-0.57} for the NER task and \textbf{0.39-0.44} for the RE task. While these results can be improved, we can interpret them as a sort of industrial baseline for the corpus. Such results can be obtained by a specialist rigorously following academic publications, but not conducting large-scale research independently.

One can easily notice the contrast between these scores and the results obtained on most often cited academic corpora such as CoNLL-2003 and SemEval2010 Task 8. In our opinion, this can be explained by domain-specific content of the corpus and by the nature of entities and relations (that are often longer and have less well-defined boundaries than ``standard'' entities). 

The last assumption can be illustrated by the fact that there is a direct correlation between the average length of the entity and the difference between token-based f-measure for entities and char-based f-measure, see Table \ref{tab:diff}.
 \begin{table}[hbt!]
    \centering
    \begin{tabular}{l|c|c|c|c|c|c|c|c|c|c|c|c}
Metrics & ACT & BIN & CMP & ECO & INST & MET & QUA & SOC\\ 
\hline
Average span-based f1 & 0.23 & 0.55 & 0.79 &  0.43 & 0.4 & 0.47 & 0.53 & 0.36\\ \hline
Average f1 diff & 0.28 & 0.03 & 0.00 & 0.23 & 0.21 & 0.27 & 0.00 & 0.19 \\ \hline
Mean no. chars & 34 & 12 & 10 & 24 & 27 & 31 & 12 & 21 \\ \hline
Mean no. tokens & 4.74 & 1.05 &  1.16 & 2.78 & 3.69 & 4.23 & 1.14 & 2.77 \\
 \end{tabular}
    \caption{ Differences in char-based f-measure and span-based}
    \label{tab:diff}
\end{table}{}

We can draw a direct comparison between RuREBus corpus and SemEval-2020 Task 11  corpus for propaganda detection \cite{DaSanMartinoSemeval20task11}. While these two corpora have completely different domains and are in different languages, both involve span extraction of long entities with sometimes less-than-clear borders and yield comparable results (0.57 f-measure for RuREBus, 0.52 for SemEval-2020 Task 11). While not all entities in industrial settings are of this type, some are, and thus, RuREBus can be treated as ``worst-case business scenario''.

\section{Conclusion} \label{sec:conclusion} 
In this paper, we deal with a real-world situation when one applies SoTA methods for NER and RE tasks. To this end, we have retrieved a large domain-specific text collection and manually annotated a small fraction of it with a `non-standard' annotations (RuREBus corpus). The BERT-based baseline, as well as other independently developed and tested models, have shown low results (f1-score \textbf{0.53-0.57} for the NER task and \textbf{0.39-0.44} for RE task). This negative result helps to learn about the extent of the gap between the academic evaluations of SoTA models and the results of the same models in practical applications. Our result is consistent with another study (in a different domain) of information extraction models (SemEval-2020 Task 11).

Indeed, our ad-hoc approach can be criticized for many reasons (e.g., for the lack of deep analysis of errors, for the lack of diverse methods, or for the presence of `non-standard' types in annotation schema). However, we argue that in industrial cases, many parameters may be less controllable than the \textit{in-vitro} setting, which leads to more laborious tasks. Thus, the RuREBus corpus can be considered as a typical ``worst-case business scenario'' for NER and RE tasks.
Future work direction include investigating domain adaptation and fine-tuning strategies and leveraging semi-supervided methods, such as cross-view training \cite{clark2018semi} to make reasonable use of unlabelled texts.

\subsubsection*{Acknowledgements.}

Work on maintenance of the annotation system, discussions of results, and manuscript preparation was carried out by Elena Tutubalina, Vladimir Ivanov, Tatiana Batura and supported by the Russian Science Foundation grant no. 20-11-20166.
Ekaterina Artemova and Veronika Sarkisyan worked on text annotation, discussions of results, and manuscript. Their work was supported by the framework of the HSE University Basic Research Program and Russian Academic Excellence Project ``5-100''. 

% \section{References}
\bibliographystyle{splncs04}
\bibliography{papers}

\begin{thebibliography}{10}
\providecommand{\url}[1]{\texttt{#1}}
\providecommand{\urlprefix}{URL }
\providecommand{\doi}[1]{https://doi.org/#1}

\bibitem{anisimovich2012}
Anisimovich, K., Druzhkin, K., Minlos, F., Petrova, M., Selegey, V., Zuev, K.:
  Syntactic and semantic parser based on abbyy compreno linguistic
  technologies. In: Computational Linguistics and Intellectual Technologies:
  Proceedings of the International Conference “Dialog” [Komp’iuternaia
  Lingvistika i Intellektual’nye Tehnologii: Trudy Mezhdunarodnoj
  Konferentsii “Dialog”]. vol.~2, pp. 90--103. Bekasovo, Russia (2012)

\bibitem{bojanowski2017enriching}
Bojanowski, P., Grave, E., Joulin, A., Mikolov, T.: Enriching word vectors with
  subword information. Transactions of the Association for Computational
  Linguistics  \textbf{5},  135--146 (2017)

\bibitem{cardellino2017low}
Cardellino, C., Teruel, M., Alemany, L.A., Villata, S.: A low-cost,
  high-coverage legal named entity recognizer, classifier and linker. In:
  Proceedings of the 16th edition of the International Conference on Articial
  Intelligence and Law. pp. 9--18 (2017)

\bibitem{carreras-marquez-2004-introduction}
Carreras, X., M{\`a}rquez, L.: Introduction to the {C}o{NLL}-2004 shared task:
  Semantic role labeling. In: Proceedings of the Eighth Conference on
  Computational Natural Language Learning ({C}o{NLL}-2004) at {HLT}-{NAACL}
  2004. pp. 89--97. Association for Computational Linguistics, Boston,
  Massachusetts, USA (2004), \url{https://www.aclweb.org/anthology/W04-2412}

\bibitem{clark2018semi}
Clark, K., Luong, M.T., Manning, C.D., Le, Q.: Semi-supervised sequence
  modeling with cross-view training. In: Proceedings of the 2018 Conference on
  Empirical Methods in Natural Language Processing. pp. 1914--1925 (2018)

\bibitem{DaSanMartinoSemeval20task11}
Da~San~Martino, G., Barr\'{o}n-Cede\~no, A., Wachsmuth, H., Petrov, R., Nakov,
  P.: {SemEval}-2020 task 11: Detection of propaganda techniques in news
  articles. In: Proceedings of the 14th International Workshop on Semantic
  Evaluation. SemEval 2020, Barcelona, Spain (September 2020)

\bibitem{devlin2019bert}
Devlin, J., Chang, M.W., Lee, K., Toutanova, K.: Bert: Pre-training of deep
  bidirectional transformers for language understanding. In: Proceedings of the
  2019 Conference of the North American Chapter of the Association for
  Computational Linguistics: Human Language Technologies, Volume 1 (Long and
  Short Papers). pp. 4171--4186 (2019)

\bibitem{dozier2010named}
Dozier, C., Kondadadi, R., Light, M., Vachher, A., Veeramachaneni, S., Wudali,
  R.: Named entity recognition and resolution in legal text. In: Semantic
  Processing of Legal Texts, pp. 27--43. Springer (2010)

\bibitem{rured}
Gordeev, D., Davletov, A., Rey, A., Akzhigitova, G., Geymbukh, G.: Relation
  extraction dataset for the russian language. In: Computational Linguistics
  and Intellectual Technologies: Proceedings of the International Conference
  “Dialog” [Komp’iuternaia Lingvistika i Intellektual’nye Tehnologii:
  Trudy Mezhdunarodnoj Konferentsii “Dialog”]. Moscow, Russia (2020)

\bibitem{hendrickx-etal-2010-semeval}
Hendrickx, I., Kim, S.N., Kozareva, Z., Nakov, P., {\'O}~S{\'e}aghdha, D.,
  Pad{\'o}, S., Pennacchiotti, M., Romano, L., Szpakowicz, S.: {S}em{E}val-2010
  task 8: Multi-way classification of semantic relations between pairs of
  nominals. In: Proceedings of the 5th International Workshop on Semantic
  Evaluation. pp. 33--38. Association for Computational Linguistics, Uppsala,
  Sweden (Jul 2010), \url{https://www.aclweb.org/anthology/S10-1006}

\bibitem{10.5555/1614049.1614064}
Hovy, E., Marcus, M., Palmer, M., Ramshaw, L., Weischedel, R.: Ontonotes: The
  90\% solution. In: Proceedings of the Human Language Technology Conference of
  the NAACL, Companion Volume: Short Papers. p. 57–60. NAACL-Short ’06,
  Association for Computational Linguistics, USA (2006)

\bibitem{huang2015community}
Huang, C.C., Lu, Z.: Community challenges in biomedical text mining over 10
  years: success, failure and the future. Briefings in bioinformatics
  \textbf{17}(1),  132--144 (2015)

\bibitem{rurebus}
Ivanin, V., Artemova, E., Batura, T., Ivanov, V., Sarkisyan, V., Tutubalina,
  E., Smurov, I.: Rurebus-2020 shared task: Russian relation extraction for
  business. In: Computational Linguistics and Intellectual Technologies:
  Proceedings of the International Conference “Dialog” [Komp’iuternaia
  Lingvistika i Intellektual’nye Tehnologii: Trudy Mezhdunarodnoj
  Konferentsii “Dialog”]. Moscow, Russia (2020)

\bibitem{joshi2020spanbert}
Joshi, M., Chen, D., Liu, Y., Weld, D.S., Zettlemoyer, L., Levy, O.: Spanbert:
  Improving pre-training by representing and predicting spans. Transactions of
  the Association for Computational Linguistics  \textbf{8},  64--77 (2020)

\bibitem{kuratov2019adaptation}
Kuratov, Y., Arkhipov, M.: Adaptation of deep bidirectional multilingual
  transformers for russian language. In: Computational Linguistics and
  Intellectual Technologies: Proceedings of the International Conference
  “Dialog” [Komp’iuternaia Lingvistika i Intellektual’nye Tehnologii:
  Trudy Mezhdunarodnoj Konferentsii “Dialog”]. pp. 333--339 (2019)

\bibitem{KutuzovKuzmenko2017}
Kutuzov, A., Kuzmenko, E.: Webvectors: A toolkit for building web interfaces
  for vector semantic models. In: Analysis of Images, Social Networks and Texts
  (AIST 2016). pp. 155--161. Springer International Publishing (2017),
  \url{http://dx.doi.org/10.1007/978-3-319-52920-2_15}

\bibitem{lample2016neural}
Lample, G., Ballesteros, M., Subramanian, S., Kawakami, K., Dyer, C.: Neural
  architectures for named entity recognition pp. 260--270 (2016)

\bibitem{leitner2019fine}
Leitner, E., Rehm, G., Moreno-Schneider, J.: Fine-grained named entity
  recognition in legal documents. In: International Conference on Semantic
  Systems. pp. 272--287. Springer (2019)

\bibitem{leitner2020dataset}
Leitner, E., Rehm, G., Moreno-Schneider, J.: A dataset of german legal
  documents for named entity recognition. arXiv preprint arXiv:2003.13016
  (2020)

\bibitem{ma2016end}
Ma, X., Hovy, E.: End-to-end sequence labeling via bi-directional
  lstm-cnns-crf. In: Proceedings of the 54th Annual Meeting of the Association
  for Computational Linguistics (Volume 1: Long Papers). pp. 1064--1074 (2016)

\bibitem{peters2018deep}
Peters, M.E., Neumann, M., Iyyer, M., Gardner, M., Clark, C., Lee, K.,
  Zettlemoyer, L.: Deep contextualized word representations. In: Proceedings of
  NAACL-HLT. pp. 2227--2237 (2018)

\bibitem{shen2017deep}
Shen, Y., Yun, H., Lipton, Z.C., Kronrod, Y., Anandkumar, A.: Deep active
  learning for named entity recognition. In: Proceedings of the 2nd Workshop on
  Representation Learning for NLP. pp. 252--256 (2017)

\bibitem{FactRuEval2016}
Starostin, A., Bocharov, V., Alexeeva, S., Bodrova, A., Chuchunkov, A.,
  Dzhumaev, S., Efimenko, I., Granovsky, D., Khoroshevsky, V., Krylova, I.,
  Nikolaeva, M., Smurov, I., Toldova, S.: Factrueval 2016: Evaluation of named
  entity recognition and fact extraction systems for russian. In: Computational
  Linguistics and Intellectual Technologies: Proceedings of the International
  Conference “Dialog” [Komp’iuternaia Lingvistika i Intellektual’nye
  Tehnologii: Trudy Mezhdunarodnoj Konferentsii “Dialog”]. pp. 702--720
  (2016)

\bibitem{stenetorp2012brat}
Stenetorp, P., Pyysalo, S., Topi{\'c}, G., Ohta, T., Ananiadou, S., Tsujii, J.:
  Brat: a web-based tool for nlp-assisted text annotation. In: Proceedings of
  the Demonstrations at the 13th Conference of the European Chapter of the
  Association for Computational Linguistics. pp. 102--107. Association for
  Computational Linguistics (2012)

\bibitem{strauss2016results}
Strauss, B., Toma, B., Ritter, A., De~Marneffe, M.C., Xu, W.: Results of the
  wnut16 named entity recognition shared task. In: Proceedings of the 2nd
  Workshop on Noisy User-generated Text (WNUT). pp. 138--144 (2016)

\bibitem{teruel2018legal}
Teruel, M., Cardellino, C., Cardellino, F., Alemany, L.A., Villata, S.: Legal
  text processing within the mirel project. In: Proceedings of the Eleventh
  International Conference on Language Resources and Evaluation (LREC 2018)
  (2018)

\bibitem{conll03}
Tjong Kim~Sang, E.F., De~Meulder, F.: Introduction to the conll-2003 shared
  task: Language-independent named entity recognition. In: Proceedings of the
  Seventh Conference on Natural Language Learning at HLT-NAACL 2003 - Volume 4.
  p. 142–147. CONLL ’03, Association for Computational Linguistics, USA
  (2003)

\bibitem{Walker}
Walker, C., Strassel, S., Medero, J., Maeda, K.: ACE 2005 Multilingual Training
  Corpus. LDC2006T06. Philadelphia: Linguistic Data Consortium (2006)

\bibitem{weber2020huner}
Weber, L., M{\"u}nchmeyer, J., Rockt{\"a}schel, T., Habibi, M., Leser, U.:
  Huner: improving biomedical ner with pretraining. Bioinformatics
  \textbf{36}(1),  295--302 (2020)

\bibitem{wu2019enriching}
Wu, S., He, Y.: Enriching pre-trained language model with entity information
  for relation classification. In: Proceedings of the 28th ACM International
  Conference on Information and Knowledge Management. pp. 2361--2364 (2019)

\bibitem{yang2017transfer}
Yang, Z., Salakhutdinov, R., Cohen, W.W.: Transfer learning for sequence
  tagging with hierarchical reccurent networks. arXiv preprint arXiv:1703.06345
   (2017)

\bibitem{zhang2017tacred}
Zhang, Y., Zhong, V., Chen, D., Angeli, G., Manning, C.D.: Position-aware
  attention and supervised data improve slot filling. In: Proceedings of the
  2017 Conference on Empirical Methods in Natural Language Processing (EMNLP
  2017). pp. 35--45 (2017),
  \url{https://nlp.stanford.edu/pubs/zhang2017tacred.pdf}

\bibitem{ZuyevK2013StatiSticalMT}
Zuev, K.A., Indenbom, M.E., Judina, M.V.: Statistical machine translation with
  linguistic language model. In: Computational Linguistics and Intellectual
  Technologies: Proceedings of the International Conference “Dialog”
  [Komp’iuternaia Lingvistika i Intellektual’nye Tehnologii: Trudy
  Mezhdunarodnoj Konferentsii “Dialog”]. vol.~2, pp. 164--172. Bekasovo,
  Russia (2013)

\end{thebibliography}

\end{document}